%% file: main.tex
\PassOptionsToPackage{table}{xcolor}
\documentclass[10pt,twocolumn,letterpaper]{article}

\usepackage[pagenumbers]{cvpr} 
\usepackage{lipsum}
\usepackage{multicol}

\newcommand{\tj}[1]{{{#1}}}

\newcommand\blfootnote[1]{%
  \begingroup
  \renewcommand\thefootnote{}\footnote{#1}%
  \addtocounter{footnote}{-1}%
  \endgroup
}

\input{preamble}

\definecolor{cvprblue}{rgb}{0.21,0.49,0.74}
\usepackage[pagebackref,breaklinks,colorlinks,citecolor=cvprblue]{hyperref}

\def\paperID{5006} 
\def\confName{CVPR}
\def\confYear{2024}

\title{Programmable Motion Generation for Open-Set Motion Control Tasks}

\author{Hanchao Liu$^{1,2*}$ \quad Xiaohang Zhan$^{2 {\dag}}$ \quad Shaoli Huang$^2$ \quad Tai-Jiang Mu$^{1 {\dag} }$ \quad Ying Shan$^2$\\
$^1$ BNRist, Tsinghua University \quad $^2$ Tencent AI Lab
}

\begin{document}

\twocolumn[{%
\renewcommand\twocolumn[1][]{#1}%
\maketitle
\begin{center}
    \centering
    \captionsetup{type=figure}
    \includegraphics[width=\linewidth]{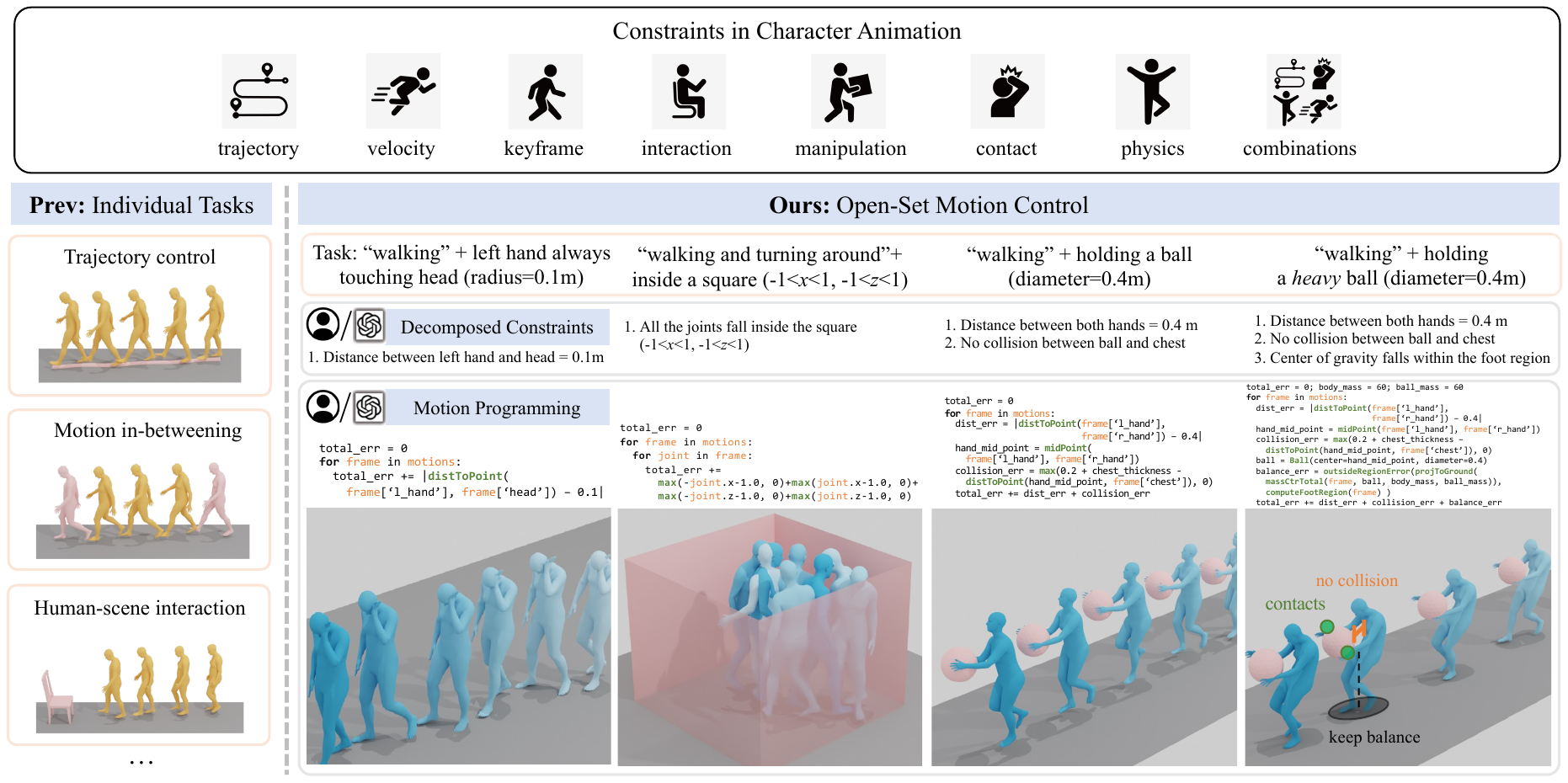}
    \captionof{figure}{We introduce \textbf{Programmable Motion Generation} as a solution for open-set human motion control. Unlike previous works that treat a finite set of motion constraints as individual tasks, we attempt to solve vast and novel tasks in a unified framework. Through Programmable Motion Generation, an arbitrary controlled motion generation task is effectively solved by simply programming an error function rather than collecting training data and designing networks. The programming is also able to be implemented automatically.
    }
    \label{fig:teaser}
\end{center}%
}]

\input{sec/0_abstract}    
\input{sec/1_intro}

\input{sec/2_related}

\input{sec/3_method}

\input{sec/4_experiment}

\input{sec/5_conclusion}

{
    \small
    \bibliographystyle{ieeenat_fullname}
    \bibliography{main}
}

\input{sec/X_supple_final}

\begin{figure*}[t]
  \centering
   \includegraphics[width=0.85\linewidth]{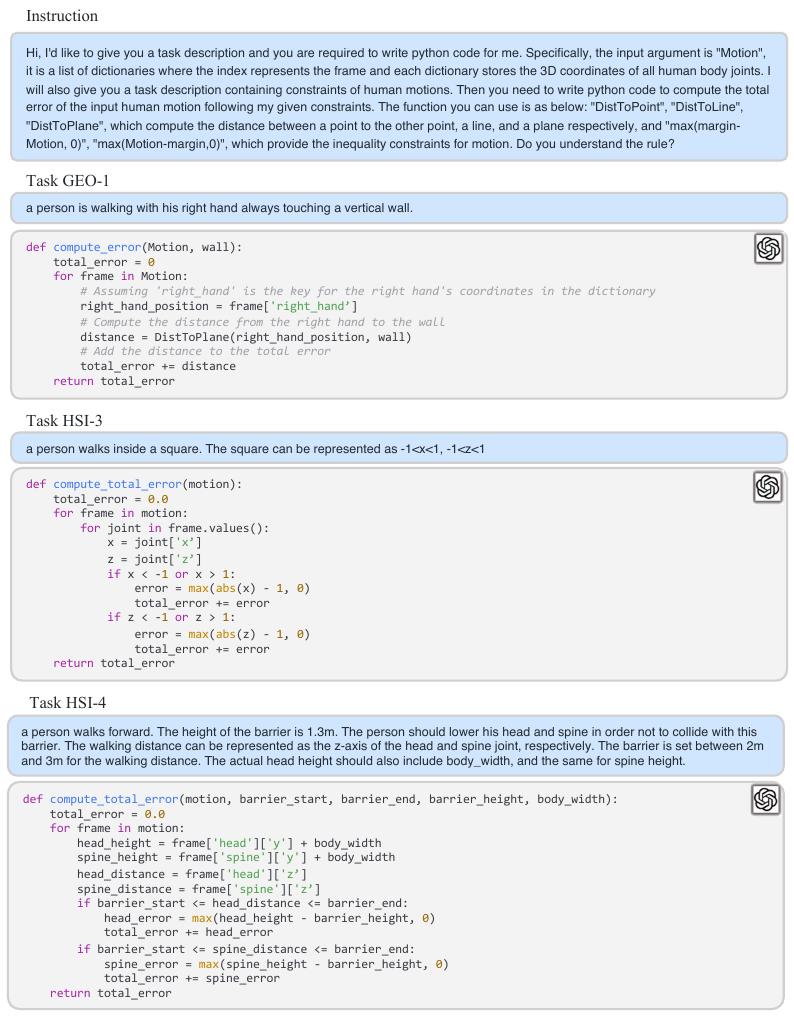}
   \caption{Motion Programming by LLM. After feeding the instruction to GPT, we provide the textual description for an arbitrary open-set motion control task. GPT will output code for the corresponding error function. We observe that GPT understands concept like \emph{touching wall} by picking the correct \emph{distToPlane} constraint, and picks correct inequality operations for tasks like \emph{avoiding overhead barrier} and \emph{walking inside a square}.  }
   \vspace{10pt}
   \label{fig:llm}
\end{figure*}


\end{document}

%% file: preamble.tex
\usepackage[dvipsnames]{xcolor}
\newcommand{\revise}[1]{{#1}}
\newcommand{\red}[1]{{#1}}


%% file: sec/0_abstract.tex
\begin{abstract}

\blfootnote{$^*$ Work done during an internship at Tencent AI Lab. }
\blfootnote{$^{\dag}$ Joint corresponding authors. {\tt xhangzhan@tencent.com, {taijiang}@tsinghua.edu.cn} }
Character animation in real-world scenarios necessitates a variety of constraints, such as trajectories, key-frames, interactions, etc. 
Existing methodologies typically treat single or a finite set of these constraint(s) as separate control tasks. These methods are often specialized, and the tasks they address are rarely extendable or customizable.
We categorize these as solutions to the close-set motion control problem. In response to the complexity of practical motion control, we propose and attempt to solve the \textbf{open-set motion control} problem. This problem is characterized by an open and fully customizable set of motion control tasks.
To address this, we introduce a new paradigm, \textbf{programmable motion generation}. 
In this paradigm, any given motion control task is broken down into a combination of atomic constraints. These constraints are then programmed into an error function that quantifies the degree to which a motion sequence adheres to them. 
We utilize a pre-trained motion generation model and optimize its latent code to minimize the error function of the generated motion.
Consequently, the generated motion not only inherits the prior of the generative model but also satisfies the requirements of the compounded constraints.
Our experiments demonstrate that our approach can generate high-quality motions when addressing a wide range of unseen tasks. These tasks encompass motion control by motion dynamics, geometric constraints, physical laws, interactions with scenes, objects or the character's own body parts, etc. All of these are achieved in a unified approach, without the need for ad-hoc paired training data collection or specialized network designs.
During the programming of novel tasks, we observed the emergence of new skills beyond those of the prior model.
With the assistance of large language models, we also achieved automatic programming. We hope that this work will pave the way for the motion control of general AI agents. 
Project page: \url{https://hanchaoliu.github.io/Prog-MoGen/}

\end{abstract}

%% file: sec/1_intro.tex
\vspace{-10pt}
\section{Introduction}
\label{sec:intro}


\begin{figure*}[t]
  \centering
   \includegraphics[width=0.95\linewidth]{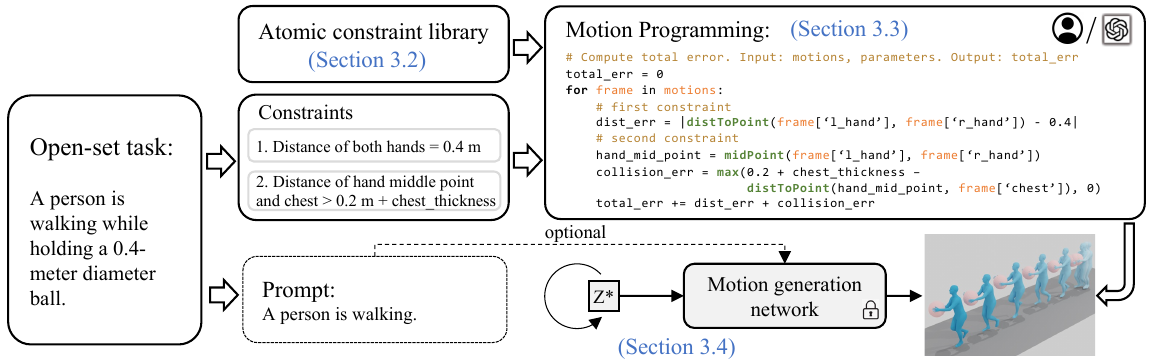}
   \caption{Overview of Programmable Motion Generation. Given an arbitrary task, we formulate it as a combination of motion constraints. Under our programming framework, by combining modules from our atomic constraint library, it is easy to program the error function to solve complex tasks just like building blocks. The programming also supports to be performed automatically by LLMs via simply providing textual descriptions of the task. Finally, \tj{the latent code $z$ of} a pre-trained motion generation network is optimized to minimize the error function, thus producing motions in high quality as well as satisfying the constraints. The prompt is optional if we use text-to-motion network as the pre-trained generative model.}
   \label{fig:overview}
   \vspace{-10pt}
\end{figure*}

Character animation techniques have extensive applications in the film and game industry, as well as in robotics \cite{nishimura2020long}. Recently, relying on large motion capture database, AI-based human motion generation methods have demonstrated their potentials when given multi-modal signals like text \cite{guo2022generating, ahuja2019language2pose, Lin_2023_CVPR, petrovich2022temos} or audio~\cite{li2021audio2gestures, alexanderson2023listen}.
However, in the practical applications of character animation, it is crucial to consider various constraints of motions, since a character is never isolated in space. 
These constraints typically include joint trajectories, motion dynamics such as velocity or acceleration, key-frames, interactions with scenes and objects, self-contacts \cite{muller2021self}, laws of physics, \textit{etc.}, and their combinations.

Artists often use Inverse Kinematics (IK) systems in Digital Content Creation (DCC) software to modify motions to meet customized constraints. However, due to the absence of motion priors, 
IK cannot ensure spatial validity among joints or temporal coherence among frames, thus usually yielding unsatisfactory results.
On the other hand, as shown in Fig.~\ref{fig:teaser}, existing AI-based animation methods typically pre-define single or a finite set of constraint(s) and formulate it as individual tasks, such as trajectory and velocity control \cite{arikan2002interactive,holden2017phase,karunratanakul2023guided, shafir2023human}, motion in-betweening \cite{harvey2020robust,qin2022motion,wei2023understanding}, human-scene/object interactions~\cite{starke2019neural,cao2020long, araujo2023circle,xu2023interdiff}, physics-based animation~\cite{peng2018deepmimic, won2020scalable,yuan2023physdiff,peng2017deeploco}, \textit{etc}.
Under such task-specific paradigm: first, for each task, the dataset and the 
methodology are specifically designed and individually trained; second, those methods intrinsically cannot deal with customized constraints or arbitrary combinations of them, thus 
\tj{being} seldom extendable or customizable. We classify those individual tasks as \emph{close-set motion control} problem.

In this paper, to confront the complexity of practical motion control, we pose a new problem, \ie \emph{open-set motion control}, where the set of motion control tasks is open and fully customizable. For example, as shown in Fig.~\ref{fig:teaser}, the generated motions of ``walking'' can be accompanied by any arbitrary 
\tj{constraint}, such as ``left hand always touching head'', ``limited in a given square'', ``holding a ball'', \textit{etc}, without special training data or network designs. To the best of our knowledge, this problem has never been solved by previous works.

To address this challenging problem, our key observations are: (1) a complicated motion control task can be broken down into several constraints; (2) almost all constraints can be measured via errors, \textit{e.g.}, using distance as an error to measure the ``contact of both hands'' constraint, and (3) the errors are mathematically additive. Based on these observations, we propose a new motion generation paradigm, \ie \emph{programmable motion generation}, \tj{where} an arbitrary \tj{controlled motion generation} task is \tj{unifiedly} solved by simply programming the error function.
Specifically, given an arbitrary motion control task, we formulate it as combinations of atomic constraints, and program them into an error function that measures how much the generated motion follows those constraints. Taking human-object interaction as an example in Fig.~\ref{fig:overview}, given a task that a person is walking while holding a 0.4 meter diameter ball, we break it down into two atomic constraints: (1) contact of hands and the ball: the distance of both hands keeps 0.4 meter; (2) avoiding collision between the ball and the chest: the distance between the mid-point of both hands and the chest joint is larger than the radius plus chest thickness. Afterwards, we program the function to compute the total error. As long as such error function is differentiable, there are many ways to optimize a pre-trained motion generation model to minimize the error. According to our statistics, almost all commonly-used constraints can be programmed as differentiable functions. In this way, the motion is optimized to satisfy the constraints while still inheriting the prior from the pre-trained generative model. 

This paradigm is extendable, \textit{e.g.}, if the ball is heavy, we can simply add another constraint to keep balance when walking, \textit{i.e.}, the ground projection point of the overall center of gravity should fall within the convex hull formed by the outline of both feet.

Additionally, to facilitate programming, we provide an atomic constraint library comprising of common atomic constraints. We also design a motion programming framework that pre-defines the input, output, as well as usable logical operations. Under the programming framework, by combining modules from the library, one can easily build complex constraints to solve customized tasks, just like building blocks. The framework and the library also make automatic programming easier. We instruct a large language model (LLM) to understand the task description and use the programming framework and the library to generate code of the error function. 
One can choose to automatically program for convenience or manually program for controllability and interpretability.

In summary, the contributions are as follows:
\begin{itemize}
    \item We pose the new problem of open-set motion control, hoping to open up new research areas for pursuing an omnipotent and generalizable intelligent agent, and providing more powerful tools for character animation developers and artists.
    \item To address the above problem, we propose programmable motion generation, a novel, flexible, customizable and versatile paradigm and its implementation.
    \item Extensive experiments show its feasibility and high motion quality for a wide range of tasks. We also observe emergence of new skills from novel tasks.
    \item Its compatibility with LLMs makes automatic execution of arbitrary open-set tasks possible, showing bigger imagination space in the future.
\end{itemize}

%% file: sec/2_related.tex
\section{Related Work}
\label{sec:related}

\noindent\textbf{Human Motion Generation.} 
Deep learning-based human motion generation has achieved great progress. Various network structures are proposed for motion generation including convolutional auto-encoder \cite{holden2016deep, hou2024causal}, variational auto-encoder (VAE) \cite{petrovich2021action}, generative adversarial network (GAN) \cite{xu2023actformer} and diffusion models \cite{tevet2022human, chen2023executing, dabral2023mofusion, zhou2023ude}. 
Apart from generating isolated human motions with text input \cite{guo2022generating, ahuja2019language2pose, Lin_2023_CVPR}, many researches focus on generating humans that interact with the surroundings and common objects \cite{wang2022towards, hassan2021stochastic, zhang2022couch, araujo2023circle, ghosh2023imos}. 
Note that these approaches usually require specific network designs for different types of conditioning signals. They are task-specific and usually incorporate task-specific domain knowledge. 
In this paper we aim to find a versatile approach that works on multiple tasks.

\noindent\textbf{Human Motion Editing and Control.}
There are also works focusing on editing or adding control to human motion generation \cite{holden2017phase,tevet2022human,shafir2023human,karunratanakul2023guided}. MDM \cite{tevet2022human} naturally supports local trajectory editing for a certain joint in a similar manner of image inpainting \cite{lugmayr2022repaint}. PriorMDM \cite{shafir2023human} extends MDM and further exploits the correlation between the edited joints and the rest of the body with an additional finetuning process to alleviate artifacts like foot skating and motion breaking. However, those inpainting-based methods only support local trajectory editing and cannot well handle global trajectories when interacting with surrounding scenes and objects. They also fail when dealing with very sparse control signals \cite{karunratanakul2023guided}. 
PFNN \cite{holden2017phase} focuses on root trajectory control but still relies on training with conditioning signals.

An alternative solution is to cast motion control as an optimization problem. Essentially inverse kinematics (IK) supports arbitrary motion editing, but it cannot guarantee high motion quality as no prior or learning is involved. \red{The recent GMD~\cite{karunratanakul2023guided} follows classifier guidance but only supports root trajectory control. The very recent OmniControl~\cite{xie2023omnicontrol} takes trajectories of arbitrary joints as control signals, but it still only receives trajectories as control signals and involves network training.} In contrast our work studies a broader and more fundamental problem by allowing any forms of constraints on arbitrary joints without re-training.

\noindent\textbf{Human Motion Priors.}
Various forms of human motion priors are proposed to help generate more plausible human poses and motions for pose estimation tasks.
Temporal consistency priors are applied on velocity and acceleration \cite{li2019estimating, zimmer2023imposing}, feature space \cite{zhang2021learning}, and DCT \cite{akhter2012bilinear}. 
Other forms of learned priors include VPoser \cite{pavlakos2019expressive}, MPoser \cite{kocabas2020vibe}, and adversarial motion priors \cite{kocabas2020vibe, peng2021amp, davydov2022adversarial}.
Recently a few motion priors are introduced for motion generation tasks. The inpainting-based editing \cite{tevet2022human} uses 
motion prior learned from the motion diffusion model (MDM). PriorMDM \cite{shafir2023human} further uses frozen MDM as a generative motion prior to generate long sequences and multi-person interactions. We also utilize pre-trained MDM 
as a strong motion prior. However, we adopt a different approach by imposing constraints and guiding it to generate motions that fit the prior.

%% file: sec/3_method.tex
\section{Programmable Motion Generation}
\label{sec:method}

\subsection{Overview}

Given an open-set motion control task, we aim to generate a motion sequence $x \in \mathbb{R}^{N \times D}$ which contains $N$ frames of $D$-dimensional poses. It is usually expressed as the rotation and position of each joint at each frame. As in Fig. \ref{fig:overview}, we first break down the task to several motion constraints and the optional condition $\mathcal{C}$. The form of $\mathcal{C}$ depends on the motion generation network we use. For example, when we use the text-to-motion network, $\mathcal{C}$ can be text prompt or left empty.
Afterwards, these constraints are programmed as an error function $F(\cdot)$ that quantifies the degree to which a motion sequence adheres to them.
We provide an atomic constraint library (Section \ref{subsection:atom_constraint}) and fundamental rules for motion programming $F$ (Section \ref{subsection:motion_programming}).
This process can be conducted manually, and we also show the potential of using LLM (\eg GPT \cite{brown2020language}) to automatically write code for $F$.

After motion programming, we formulate this motion control task as an optimization problem:
\begin{equation}
 \min_{z} ~F(G_{\theta}(z, \mathcal{C}), p),
 \label{eq:opt}
\end{equation}
where $\theta$ is the frozen weight of a motion generation model $G_\theta$ and $p$ is the parameters affiliated to this task. Our goal is to optimize the latent vector $z$ for the generative model so that the generated motion sample $x=G_{\theta}(z, \mathcal{C})$ adheres to those constraints.
We present the solution for this optimization problem in Section \ref{subsection:latent_opt}.

\subsection{Atomic Constraints}
\label{subsection:atom_constraint}
Theoretically, the total error function $F$ can be composed of any error $E(x)$ that is differentiable with respect to $x$.
Here we introduce an atomic constraint library in a modular and systematic way to support various tasks. They are representative spatial and temporal constraints that serve as building blocks for the error function $F$. For convenience, we denote the motion of $j$-th joint as $x_j$, the position of $j$-th joint in the global coordinate system as $x^{pos}_j = T\left(x_j\right)$, where $T$ transforms the motion $x_j$ to global joint positions and it is differentiable.

\noindent\textbf{\red{Absolute Position Constraint}}
 requires the trajectory $x^{pos}_j$ of $j$-th joint to be close to a given trajectory $\hat x^{pos}_j$ and is in the form of \emph{L-n} norms, \ie, $E(x^{pos}_j, \hat x^{pos}_j) = |x^{pos}_j - \hat x^{pos}_j|_n$. Existing trajectory-based motion control tasks \cite{shafir2023human,karunratanakul2023guided, xie2023omnicontrol} constitute a subset of this constraint.
It can also serve as a regularization term if we do not wish to change too much from the motion generated by original $G_{\theta}$.

\noindent\textbf{High-order Dynamics Constraint} constrains motion dynamics of joints instead of positions. A typical example is to constrain the magnitude and orientation of velocity or acceleration for certain joints. This constraint is in the form of $E(x_j^{(k)}, \hat x_j^{(k)})$ by taking the $k$-th numerical differential of $x_j$ and $\hat x_j$.

\begin{figure}[t]
  \centering
\includegraphics[width=\linewidth]{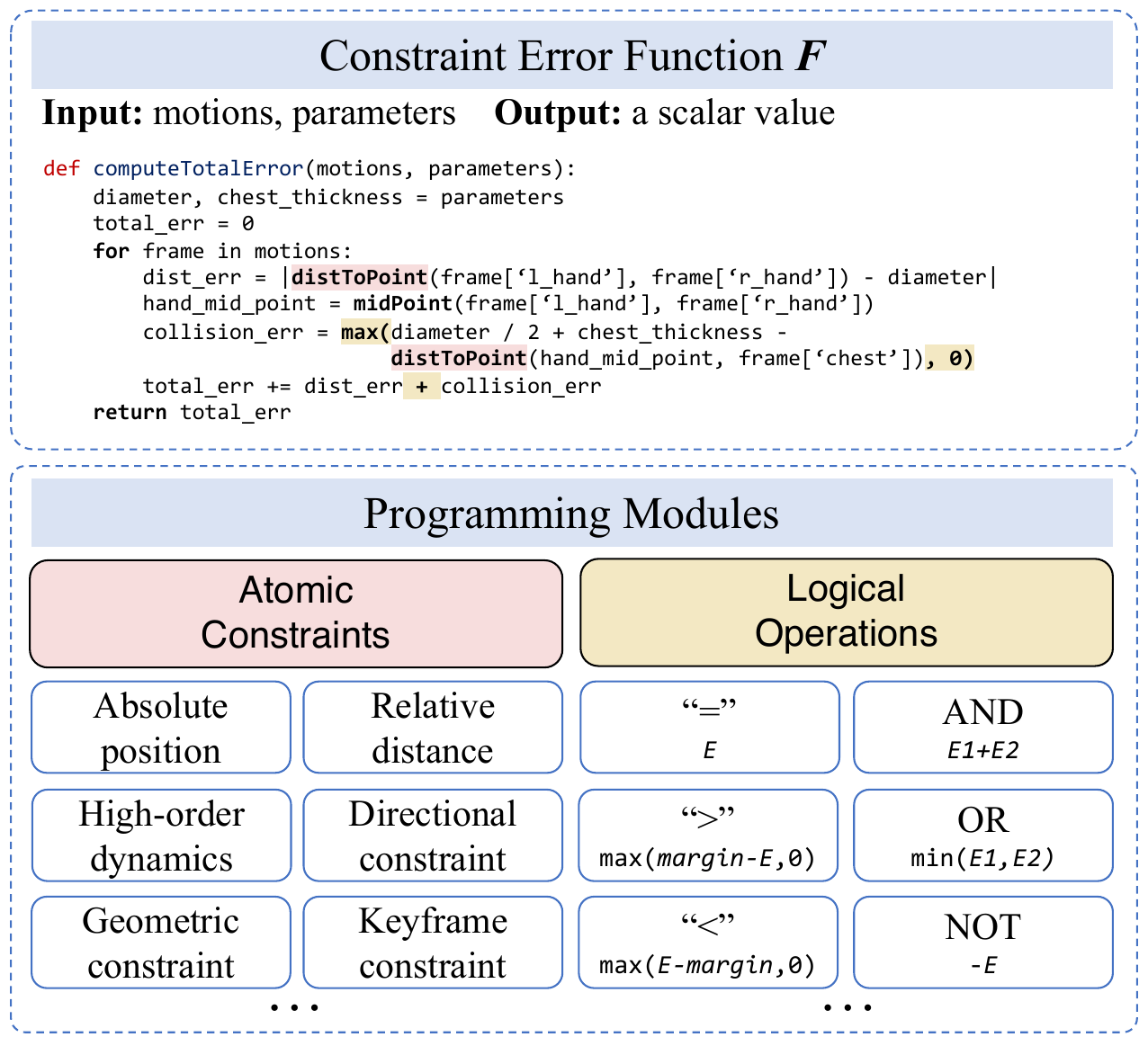}
   \caption{The programming framework that pre-defines the input, output, atomic constraints and the redesigned logical operations as building blocks for motion programming. The example code corresponds to the task of ``holding a ball''.}
   \label{fig:err_template}
\end{figure}

\noindent\textbf{Geometric Constraints} constrain a joint $x^{pos}_j$ on a geometric primitive $P$ in the global coordinate system, such as a curve or a surface, denoted by $E(x^{pos}_j, P)$. As common cases, we implement \emph{distToLine}, \emph{distToPlane}, \etc in our constraint library. Note that constraining a joint on a line differs from the aforementioned point-wise trajectory constraint, and the latter is stricter than the former.

\noindent\textbf{Relative Distance Constraint} models relationships between two joints, \eg, the distance of any two joints is denoted by $E(x^{pos}_j, x^{pos}_k)$. Similarly, the angle between two joints also belongs to this category.

\noindent\textbf{Directional Constraint} requires a bone consisting of $x_j$ and its parent joint $parent(x_j)$ to point at a given direction $d$, denoted by $E\left(x^{pos}_j - parent(x^{pos}_j), d\right)$.

\noindent\textbf{Key-frame Constraint} enforces constraint at certain timestamps. For this purpose, we can define the aforementioned constraints at some certain timestamps $t$ only, in the form of $E\left(E_{\text{spatial}}\left(x, *\right),t\right)$, where $E_{\text{spatial}}$ is any constraint irrelevant to time.

One can always write customized constraints to extend the library if necessary. For example, if we want the agent to maintain body balance when performing a certain task, \textbf{Centor-of-mass Constraint} is required. It means the ground projection point of the overall center of gravity should fall within the convex hull formed by the outline of both feet. It is quite extendable by using your imagination. For example, what if the agent is subjected to some additional external forces while maintaining balance, such as pull force or centrifugal force?

\subsection{Motion Programming} 
\label{subsection:motion_programming}
To further facilitate programming, we provide a motion programming framework consisting of the following rules.

\noindent\textbf{Input and output}. The input consists of ``motions'' and ``parameters''. The ``motions'' is a list of dictionaries containing information of joints. The ``parameters'' includes task-related constants. The output is a scalar value representing the total error.

\noindent\textbf{Logical operations}. We redesign some of the logical operations in standard programming language to better support motion programming.
\begin{itemize}
    \item ``$>$'' implemented by $max(margin-E, 0)$, means the error should be larger than a given margin. It is commonly used in obstacle avoidance.
    \item ``$<$'' implemented by $max(E-margin, 0)$, means the error should be less than a given margin.
    \item ``AND'' implemented by $E_1+E_2$, means both constraints are satisfied.
    \item ``OR'' implemented by $min(E_1, E_2)$, means one of the constraints is satisfied.
    \item ``NOT'' implemented by $-E$, means the error should be as large as possible. It is used to keep the agent as far away as possible from some geometric objects.
\end{itemize}

\noindent\textbf{Other programming rules}. Conditions like ``if-elif-else'' and loops like ``for'' are supported. It means we allow the constraints to be triggered by some customized conditions, and repeatedly applied to different frames and joints. At last, the error function is required to be differentiable to the input motion.

A template of the error function is shown in Fig. \ref{fig:err_template}.

\subsection{Latent Noise Optimization}
\label{subsection:latent_opt}

As for the optimization in Eq. (\ref{eq:opt}), we utilize a pre-trained motion diffusion model (MDM) \cite{tevet2022human} in our experiments as the prior model.
Specifically, we adapt MDM to its DDIM \cite{song2020denoising} form so that the latent noise $z$ is a single vector. We use Adam~\cite{kingma2014adam} as the optimizer in all the experiments, though other optimizers such as L-BFGS are also supported.

The human motion has invariance in translation and rotation on the horizontal plane. For tasks with constraints related to horizontal positions or rotations, we can relax the constraint by transforming it to an equivalent constraint using spatial transformation. This reduces the difficulty for the original optimization problem. For example, the constraint ``touching a vertical plane whose equation is $z=10$'' is firstly transformed to ``touching a vertical plane whose equation is $z=0$''; after optimization, the motion is then transformed back to satisfy the original constraint.

\setlength{\tabcolsep}{6pt}
\begin{table*}[t]
    \centering
    \small
    \begin{tabular}{l|cc|cc|ccc} 
    \toprule
     \multicolumn{8}{c}{Task HSI-1: head height constraint} \\
    \toprule
   Method & Foot Skate $\downarrow$  & Max Acc. $\downarrow$ &  C.Err. $\downarrow$ & Unsucc. Rate $\downarrow$  & FID $\downarrow$ &  Diversity $\rightarrow$ & R-prec. (Top3) $\uparrow$\\  
   \midrule
   MDM (Unconstrained) \cite{tevet2022human}    & 0.086 & 0.097 & 0.118 & 0.718 & 0.545 & 9.656 & 0.610 \\ \midrule
   MDM Edit \cite{tevet2022human} & 0.094 & 0.148 & \cellcolor{red!10}0.109 & \cellcolor{red!10}0.645 & 0.554 & 9.656 & 0.614  \\
   IK       & 0.093 & \cellcolor{red!10}0.414 & 0.012 & 0.088 & 0.545 & 9.653  & 0.610\\
   IK+Reg.  & \cellcolor{red!10}0.269 & 0.121 & 0.012 & 0.088 & 0.782 & 9.509  & 0.603\\   \midrule
   Ours     & 0.075 & 0.094 & 0.012 & 0.088 & 0.556 & 9.611   & 0.597 \\ 
  \bottomrule
    \end{tabular}
    \caption{Comparison with other methods with constraints sampled from groundtruth HumanML3D test set. The constraints are imposed on the first, central and last frames. MDM (Unconstrained) serves as a numerical reference. The failure of any single indicator (marked in red) means the failure of the entire task. \red{Baseline methods always fail in certain metrics while ours performs generally well on all metrics. }}
    \label{tab:1}
\end{table*}
\setlength{\tabcolsep}{1.4pt}

\setlength{\tabcolsep}{11.2pt}
\begin{table*}[t]
    \centering
    \small
    \begin{tabular}{l|ccc|ccc} 
    \toprule
    & \multicolumn{3}{c|}{\red{Task HSI-2: avoiding barrier}} &  \multicolumn{3}{c}{Task HSI-3: walking inside a square}  \\
    \midrule
   Method & Foot Skate $\downarrow$  & Max Acc. $\downarrow$ &  C.Err. $\downarrow$ & Foot Skate $\downarrow$  & Max Acc. $\downarrow$ &  C.Err. $\downarrow$  \\  
   \midrule
   MDM (Unconstrained) \cite{tevet2022human} & 0.096 & 0.126 & 0.454 & 0.096 & 0.126 & 0.301  \\ \midrule
   IK              & 0.132 & \cellcolor{red!10}1.919 & 0.047 & 0.139 & \cellcolor{red!10}0.292 & 0.015  \\
   IK+Reg.         & \cellcolor{red!10}0.589 & 0.361 & 0.047 & \cellcolor{red!10}0.215 & 0.128 & 0.015   \\   \midrule
   Ours            & 0.189 & 0.150 & 0.097 & 0.125 & 0.093 & 0.012   \\ 
  \bottomrule
  \toprule
    & \multicolumn{3}{c|}{Task GEO-1: hand touching wall} &  \multicolumn{3}{c}{Task HOI-1: moving object}  \\
    \midrule
   Method & Foot Skate $\downarrow$  & Max Acc. $\downarrow$ &  C.Err. $\downarrow$ & Foot Skate $\downarrow$  & Max Acc. $\downarrow$ &  C.Err. $\downarrow$ \\  
   \midrule
   MDM (Unconstrained) \cite{tevet2022human}   & 0.096 & 0.126 & 0.233 & 0.029 & 0.026 & 1.701  \\ \midrule
   MDM Edit \cite{tevet2022human}         & 0.161 & 0.147 & \cellcolor{red!10}0.141 & 0.029 & 0.032 & \cellcolor{red!10}1.739   \\
   PriorMDM \cite{shafir2023human}         & \cellcolor{red!10}0.350 & 0.197 & \cellcolor{red!10}0.185 & \cellcolor{red!10}0.327 & 0.213 & \cellcolor{red!10}1.884     \\
   IK                & 0.147 & 0.187 & 0.010 & \cellcolor{red!10}0.408 & \cellcolor{red!10}0.919 & 0.011  \\
   IK+Reg.           & \cellcolor{red!10}0.536 & 0.117 & 0.010 & \cellcolor{red!10}0.405 & 0.037 & 0.011 \\   \midrule
   Ours              & 0.110 & 0.104 & 0.023 & 0.114 & 0.068 & 0.028  \\ 
  \bottomrule
    \end{tabular}
    \caption{Comparison with other methods on unseen tasks. MDM Edit and PriorMDM cannot address these tasks natively. We adapt them with ad-hoc tricks to fit these tasks. MDM (Unconstrained) serves as a numerical reference. The failure of any single indicator (marked in red) means the failure of the entire task. \red{Baseline methods always fail in certain metrics while ours achieves good balance on motion quality and reaching the given constraints.} }
    \label{tab:2}
\end{table*}
\setlength{\tabcolsep}{1.4pt}

\section{Task and Applications}
\label{section:task}
In this section, we show how to combine atomic constraints to constitute a wide range of open-set motion control tasks and applications. For each task category we present several specific sub-tasks for the later evaluation.

\subsection{Motion Control with High-order Dynamics}
The tasks related to velocity or acceleration can be solved via high-order dynamics constraints. We conducted the following specific task in our experiments:

\noindent\textbf{Task HOD-1:} specifying the velocity (both magnitude and orientation) for several key-frames. This task uses ``high-order dynamics constraint'' and ``key-frame constraint''.

\subsection{Motion Control with Geometric Constraints}
Geometric constraints are common in the real world such as \emph{hand touching a wall}, \emph{feet on a balance beam}. These tasks are supported by calling \emph{geometric constraints}. They are significantly different from trajectory control tasks which are required to specify the exact joint positions at each timestamp. Geometric constraints, as looser constraints, are more suitable for such tasks like \emph{hand touching a wall} that do not need to pre-define the trajectories.
Note that the constraint relaxation strategy can be applied in these tasks. The representative tasks in our experiments include:

\noindent\textbf{Task GEO-1:} walking with hand touching a vertical wall.

\noindent\textbf{Task GEO-2:} walking with feet on a balance beam.

\subsection{Human-Scene Interaction} 
Tasks related to human-scene interactions can be solved by combining multiple constraints and logical operations. The representative tasks conducted in the experiments include:

\noindent\textbf{Task HSI-1:} constraining the head heights on the first, central and last frames. 
This task uses \red{``geometric constraint''} and ``key-frame constraint''.

\noindent\textbf{Task HSI-2:} \red{head} avoiding an overhead barrier on a specified key-frame. This task uses \red{``geometric constraint'', ``$<$ operation'', and ``key-frame constraint''}.

\noindent\textbf{Task HSI-3:} constraining a human to walk inside a square area. This task uses \red{``geometric constraint'', ``$<$ operation'' and ``$>$ operation''}.

\noindent\textbf{Task HSI-4:} avoiding an overhead barrier specified by its position on the z-axis. This task uses ``geometric constraint'' and ``$<$ operation''.

\noindent\textbf{Task HSI-5:} constraining a human to walk in a narrow gap between two walls specified by the x-axis. This task uses \red{``geometric constraint'', ``$<$ operation'' and ``$>$ operation''}.

\subsection{Human-Object Interaction}
Humans usually interact with objects by hands in actions like \emph{holding, carrying} and some other body parts like hips in actions like \emph{sitting}. These tasks can be solved via combinations of constraints and logical operations. The representative tasks in our experiments include:

\noindent\textbf{Task HOI-1:} moving an object from one place to another. Both starting and end positions for \red{the controlled hand are specified}. This task uses ``absolute position constraint'' and ``key-frame constraint''.

\noindent\textbf{Task HOI-2:} carrying a large ball with its diameter specified. This task uses ``relative distance constraint'' and ``$>$ operation''.

\subsection{Human Self-Contact}
Moreover, we handle human self-contact by applying \emph{relative distance constraint} on those joints that are in contact with each other. The task in our experiment is:

\noindent\textbf{Task HSC-1:} walking with a hand always touching the head. This task uses ``relative distance constraint''.

\subsection{Physics-based Generation} Lastly, our framework supports complex physics-based generation. For example, given the mass of each bone for a body and using \emph{center-of-mass constraint}, we can generate physically plausible motions that conform to the physical law of gravity. The tasks conducted in our experiments are:

\noindent\textbf{PBG-1:} standing with single foot and keep balanced. This task uses ``absolute position constraint'' and ``center-of-mass constraint''.

\noindent\textbf{PBG-2:} carrying a heavy ball and keeping balanced at the same time. This task uses ``relative distance constraint'', ``center-of-mass constraint'' and ``$>$ operation''.

%% file: sec/4_experiment.tex
\section{Experiments}
\label{sec:exp}

\begin{figure*}[t]
  \centering
   \includegraphics[width=\linewidth]{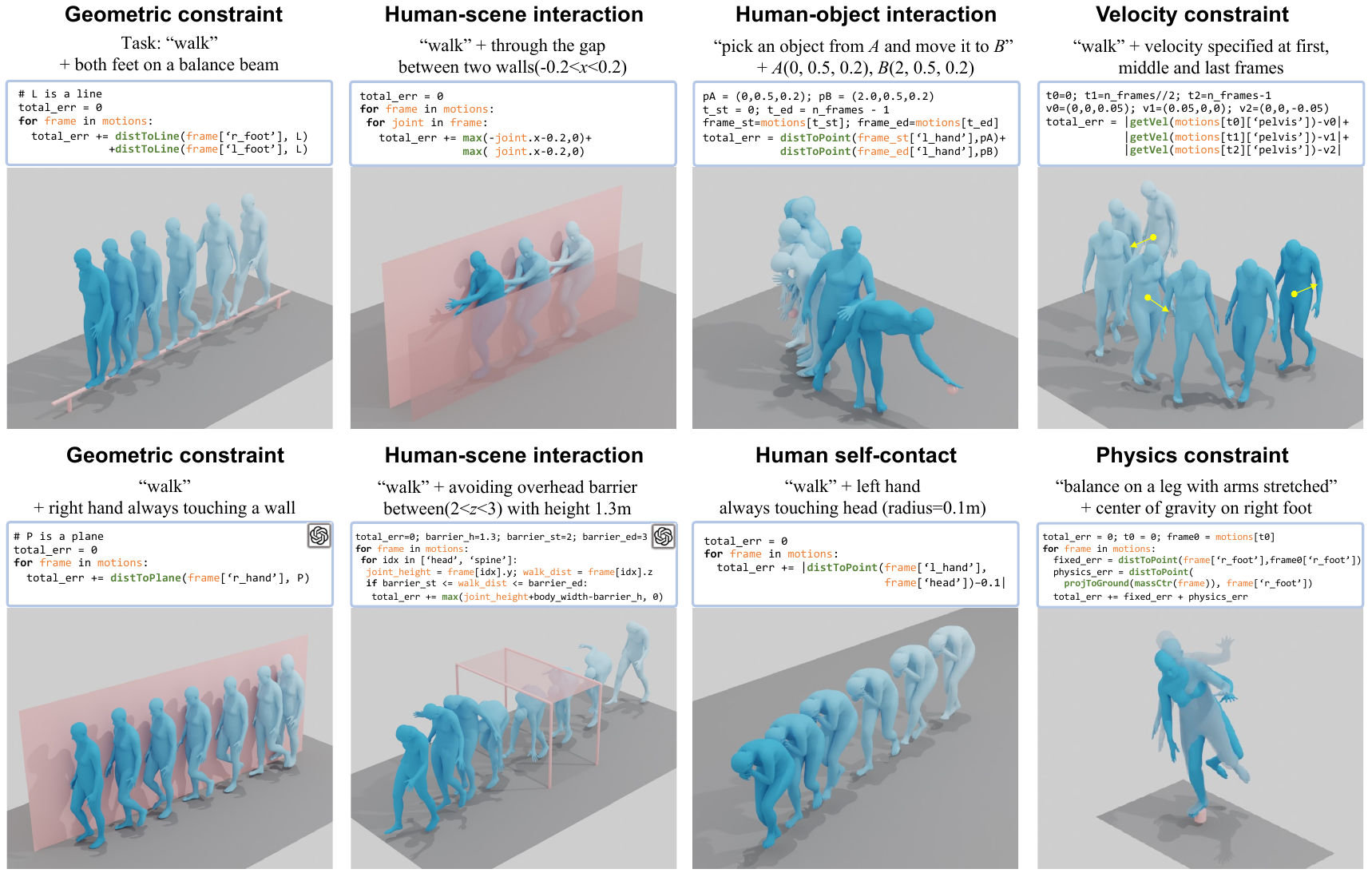}
   \caption{Qualitative examples of our method for diverse open-set motion control tasks. \red{The task, error function code and generated motion are demonstrated for each example. The code labeled with GPT marker is generated by GPT given the task description in text.}}
   \vspace{-3pt}
   \label{fig:main_demo}
\end{figure*}

As our open-set motion control problem deviates from standard text-to-motion generation \cite{guo2022generating} and trajectory-based motion control \cite{shafir2023human}, we evaluate our method on a set of pre-defined sub-tasks defined in Section \ref{section:task}. Details for each sub-task are provided in the supplementary material.

\subsection{Evaluation Metrics}

For measuring non-semantic motion quality, we use \textbf{foot skating ratio (Foot Skate)} proposed in \cite{karunratanakul2023guided} to measure the motion coherence and over-smoothing artifacts, and use \textbf{maximum joint acceleration (Max Acc.)} $\max \{ \ddot{x}^{pos}_i \}$ in a generated sample to measure frame-wise inconsistency. For semantic-related motion quality, we adopt commonly-used \textbf{Frechet Inception Distance (FID)}, Diversity and R-Precision as in \cite{shafir2023human}. Moreover, we use \textbf{constraint error (C. Err)} in MAE to measure how well the generated motion satisfies the given constraints. \red{The unsuccess rate is defined as the percentage of the generated samples which fail to meet all the constraints within 5 cm threshold.}
Note that the semantic-related metrics require that the imposed constraints also come from the groundtruth data distribution. Therefore, for unseen constraints we only evaluate on non-semantic motion quality metrics and constraint errors.

\subsection{Baselines}
We compare our method with several baseline methods. (1) \textbf{Inverse Kinematics (IK)}. The optimization process is performed on the motion \red{$x$} instead of backpropagating to the latent noise $z$. (2) \textbf{Inverse Kinematics with regularization (IK+Reg.)}. The L2-norm regularization \red{$|x_{[i+1]} - x_{[i]}|_2$} is added to help alleviate the frame inconsistency. (3) \textbf{Motion editing of Motion Diffusion Model (MDM Edit)} \cite{tevet2022human}. We first use MDM to generate trajectories for both root joint and controlled joint that meet the given constraint and then perform inpainting using these trajectories. However, as retrieving joint positions directly leads to invalid bone lengths, we choose to recover the final result from joint rotations with a skeleton template. (4) \textbf{PriorMDM finetuned control} \cite{shafir2023human}. It builds on MDM Edit and further finetunes the model parameters to capture the relationship between the clean controlled joint and the remaining joints.

\subsection{Implementation Details}
We use the official weight of MDM \cite{tevet2022human} pre-trained on HumanML3D \cite{guo2022generating} and keep it frozen. We use its DDIM version with \tj{a step of} $T_{\text{MDM}}=100$,
which makes our latent noise optimization faster. For a fair comparison, all the baseline methods also use the same DDIM model. 
\red{We find that optimizing with learning rate 0.005 and 100 optimization steps generally works well for a majority of tasks. }
More details are provided in the supplementary material.

\subsection{Results and Evaluation}

\noindent\textbf{Quantitative Evaluation.}
\red{We evaluate on tasks with both \emph{known} constraints (Table~\ref{tab:1}) and \emph{unseen} constraints (Table~\ref{tab:2}).}
As in Table~\ref{tab:1}, we show high-quality and coherent motion over baselines including IK and MDM Edit methods, which always fail in some certain metrics (marked in red background in the table). Similarly, comprehensive evaluation on four unseen sub-tasks (Table~\ref{tab:2}) shows that our method achieves good balance between motion quality and constraint errors. Especially, IK produces inconsistent motion (failed in Max. Acc.) when the added constraints are sparse, and generates over-smooth motion (failed in Foot Skate) if imposing regularization terms for frame consistency. \red{Inpainting methods are not able to produce motions that are faithfully constrained.}

\noindent\textbf{Qualitative Evaluation.} In Fig. \ref{fig:main_demo}, we demonstrate the versatility of our approach by solving a series of open-set tasks described in Sec. \ref{section:task}. 
Our method generates high quality and visually coherent motions under various constraints.
Moreover, our method performs well for tasks with both single and complicated multiple constraints. 
Especially, inpainting-based methods are unable to deal with inequality constraints and those constraints in which all body joints need to be edited, 
such as center-of-mass constraint.

\noindent\textbf{Motion Control for Unseen Tasks.}
If we construct a set of unseen constraints that are new to the generation model, our method is still able to generate quite reasonable actions. 
For example, for ``walking between two walls'', the arms are brought together and the shoulders are shrank to adapt to the narrow space. This suggests that the proposed approach intriguingly demonstrates a certain level of proficiency in fostering the emergence of new skills for motion generation.

\noindent\textbf{Motion Programming by LLM.}
Apart from manually programming the task into constraints, in Fig. \ref{fig:main_demo} we show the potential for an LLM with reasoning ability to translate task description into constraints and code the error function $F$, which is similar to \cite{gupta2023visual, xu2023creative}. 
We observe that GPT understands concept like \emph{touching wall} by picking the correct \emph{distToPlane} constraint, and picks correct inequality operations for tasks like \emph{avoiding overhead barrier} and \emph{walking inside a square}. 
More evaluation is in the supplementary.

\subsection{Analysis}

\begin{figure}[t]
  \centering
   \includegraphics[width=\linewidth]{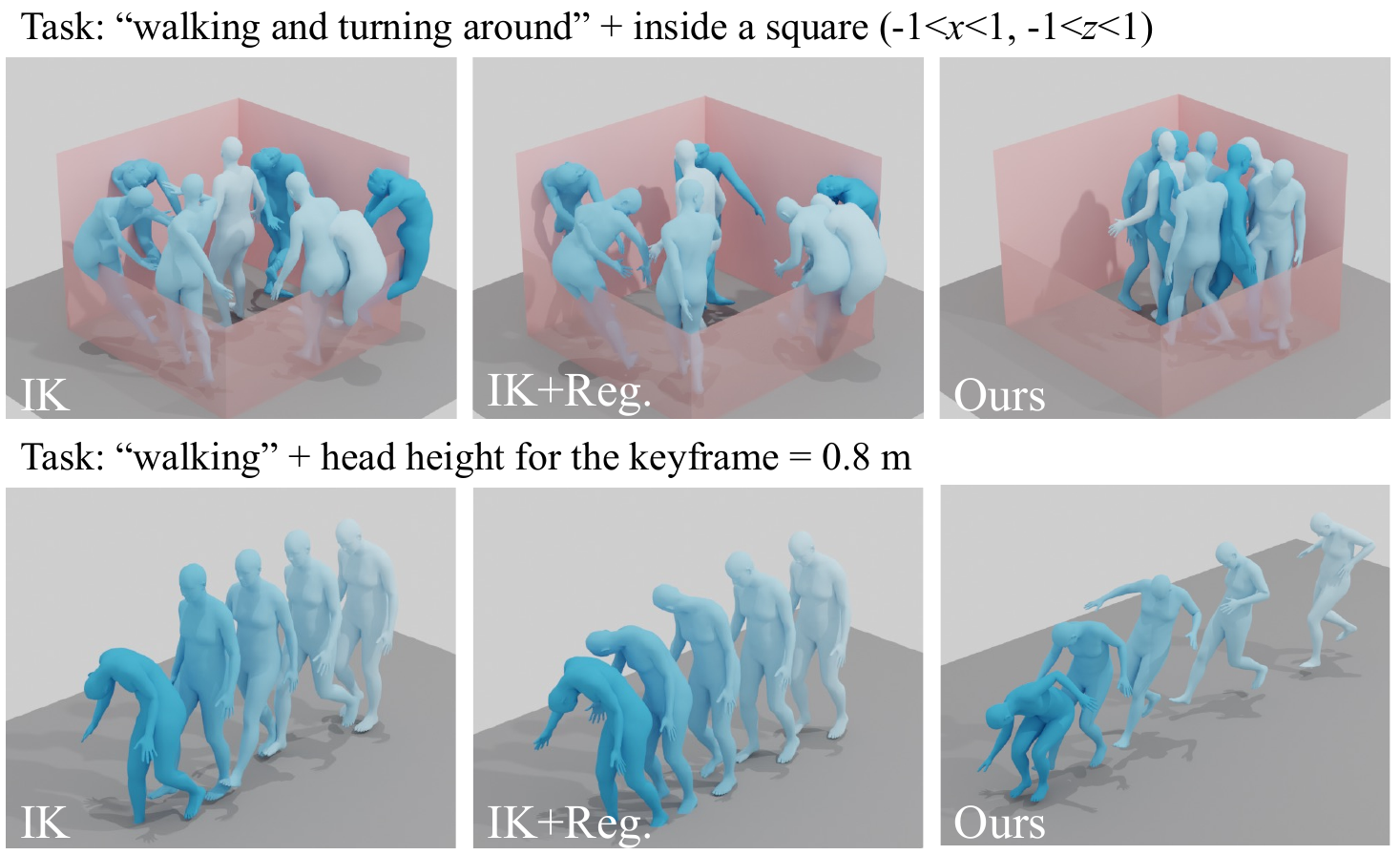}
   \caption{Effect of our motion prior. \red{Top row: Ours generates valid poses while IK and IK+Reg produce invalid ones. Bottom row: IK generates incoherent motion and IK+Reg generates over-smooth motion with massive foot skating. Our method generates coherent motion while adhering to the given constraint.} }
   \label{fig:compare}
\end{figure}

\textbf{Effect of motion prior.} As in Fig. \ref{fig:compare}, in the task of \emph{walking inside a square}, our method generates valid poses while IK and IK+Reg. produce invalid ones. 
Moreover, this type of whole-body inequality constraint cannot be handled by inpainting-based methods like MDM Edit and  PriorMDM.
In the task of \emph{head height constraint}, IK generates incoherent motion, and IK+Reg. generates over-smooth motion with massive foot skating. Our method generates coherent motion while adhering to the given constraint.

\setlength{\tabcolsep}{11pt}
\begin{table}[t]
    \centering
    \small
    \begin{tabular}{lc} 
    \toprule
   Method & Bone Length Incorrect Ratio  \\  
   \midrule
   MDM (Unconstrained)   & 0.048     \\
   MDM Edit (Position)    & 0.525 \\ 
   Ours         & 0.051\\ 
  \bottomrule
    \end{tabular}
    \caption{Comparison of effect on bone length preservation in the task \emph{head height constraint}. The inpainting-based method fails to preserve correct bone lengths if recovering from local joint positions. Ours well preserves bone lengths for the generated motions.}
    \label{tab:bone_length}
    \vspace{-13pt}
\end{table}
\setlength{\tabcolsep}{1.4pt}

To show the effect of bone length preserving, 
we further analyze the correctness of neck lengths in the generated motions for the task \emph{head height constraint} in Table \ref{tab:1}. As shown in Table \ref{tab:bone_length}, we can preserve bone lengths even if we \red{recover from local joint positions}.
The inpainting-based method MDM Edit struggles with local joint positions converted from global trajectories. 
The denoising process cannot remedy sparse and invalid inpainting signals, therefore generating motions with invalid bone lengths.

%% file: sec/5_conclusion.tex
\section{Conclusion}
\label{sec:conclu}

In this work, we present the new problem of open-set motion control.
We propose a new paradigm for this problem, namely programmable motion generation. The key idea is to formulate an arbitrary task as an error function built from atomic constraints and logical operations and use it to guide a pre-trained motion generation model to generate motion that meets these constraints. 
In the future work, we will extend the current framework to whole-body generation which allows more details, and study how to enable automatic constraint generation in large and rich semantic scenes.

{\footnotesize \noindent\textbf{Acknowledgements} This work was supported by the National Science and Technology Major Project (2021ZD0112902), the National Natural Science Foundation
of China (62220106003), and the Research Grant
of Beijing Higher Institution Engineering Research Center
and Tsinghua-Tencent Joint Laboratory for Internet Innovation Technology.}

%% file: sec/X_supple_final.tex
\clearpage
\maketitle

\setcounter{figure}{0}
\renewcommand{\thefigure}{B\arabic{figure}}

\setcounter{table}{0}
\renewcommand{\thetable}{A\arabic{table}}


\section*{A. Experiment Details}
\label{sec:rationale}

\subsection*{A.1 Tasks for Quantitative Evaluation}
We design two evaluation protocols for sub-tasks in Section 4. The first protocol is \emph{task with known constraints}, which means the added constraints are sampled from existing human motion datasets, \textit{i.e.}, HumanML3D test set \cite{guo2022generating} in our experiments. In this way, in addition to non-semantic motion quality metrics and constraint errors, we can evaluate on semantic-related motion quality as well since we have groundtruth motions. The second protocol is \emph{task with unseen constraints}, which means the added constraints do not come from existing motions and are designed by ourselves to evaluate the generation capability on real open-set motion control tasks. We experiment on \red{one sub-task} for known constraints in Table 1 in the main text, and four sub-tasks for unseen constraints in Table 2 in the main text. 

\noindent\textbf{Task with known constraints.} For Task HSI-1 \emph{head height constraint} in Table 1, we constrain the head height for three specified key-frames, \textit{i.e.}, first, middle and last frames to be equal to that in the motions sampled from HumanML3D test set. The text prompt and motion length for generation are also obtained from that motion sample. 
The constraint error for evaluation is the mean absolute error (MAE) averaged over the three key-frames. 
We follow PriorMDM \cite{shafir2023human} for evaluating metrics including FID, R-precision and Diversity, and follow GMD \cite{karunratanakul2023guided} for evaluating Foot skating ratio. 
The quantitative evaluation is conducted on 544 generated samples. 

\noindent\textbf{Task with unseen constraints.} In Table 2, for \red{Task HSI-2 \emph{avoiding overhead barrier}}, we constrain the head height to be lower than 0.5 m for the middle frame and higher than 1.5 m for the first and last frames to ensure normal standing poses at the beginning and the end. We also constrain the heights for both feet \revise{to be close to the ground.} 
 \revise{Note that this is a challenging task due to the low head height, and 
the combined constraints prevent trivial generations like \emph{stepping on stairs} or \emph{always lying on the ground}.}
The constraint error for evaluation is defined as MAE for the head height and foot heights.

For Task HSI-3 \red{\emph{walking inside a square}}, we constrain the walkable area to be a square $-1<x<1, -1<z<1$. 
\red{The constraint error for evaluation is defined as the per-joint MAE averaged over x- and z-axis and all frames.}
\begin{align}
    \text{Err}(x) &= \frac{1}{4 N N_j} \sum\limits_{t=1}^N \sum\limits_{j=1}^{N_j} \sum\limits_{dim \in \mathcal{D}}  \\ \nonumber
      \max(&-x^{pos}_{j,t,dim}-1,0) + \max(x^{pos}_{j,t,dim}-1,0)
\end{align}
where $\mathcal{D}$ includes x-axis and z-axis and $N_j$ is the number of joints.

For Task GEO-1 \red{\emph{hand touching wall}}, we constrain the left hand (joint 20) always on a vertical plane. The plane is randomly sampled with its distance to the origin no greater than 3.
The constraint error for evaluation is defined as the mean distance between the controlled hand and the given plane averaged over all frames.

For Task HOI-1 \emph{moving object}, we constrain on the global positions of the left hand (joint 20) at the first and last key-frames. We specify a set of beginning and end hand positions. 
The constraint error for evaluation is defined as the \red{mean distance between the hand and the goal averaged over the two key-frames.}

For the first three tasks, \textit{i.e.}, Task HSI-2, HSI-3 and GEO-1, the text prompts and motion lengths are sampled from a selected set of samples from HumanML3D test set, mainly involving the action \emph{walking}. \red{The sample ids are listed below: 000130,
000178,
000285,
000337,
000363,
000600,
000665,
000679,
000759,
000998,
000099,
000696,
000700,
003703,
001161,
001617,
001848,
003193,
003437,
004455 and their mirrored ones.}
For Task HOI-1, we manually compose a set of text prompts related to action
\emph{moving} such as \emph{a person moves an object from a place to another place}. The quantitative evaluation for each unseen task is conducted on running 32 generated samples.

\subsection*{A.2 Baseline Details}

\noindent\textbf{Unconstrained MDM.} The original motion representation for MDM \cite{tevet2022human} contains both local joint positions and joint rotations. For simplicity we recover global joint positions (joint positions in the global coordinate) from local joint positions.
The unconstrained MDM only serves as a numerical reference.

\noindent\textbf{IK and IK+Reg.}
We implement IK as an ablated version of our method, in which the gradient $\nabla F$ is backpropagated to motion $x$ instead of the latent vector $z$. We also consider a variant IK with regularization (IK+Reg.), in which we add a L2-norm regularization term on all joints \red{$L_{reg}=|x_{[i+1]} - x_{[i]}|$}, where $i$ is the temporal index. This results in a combined error function $L_{constraint}+w L_{reg}$. We empirically set the regularization weight $w=1.0$. 
We obtain global positions from joint rotations with a human skeleton template with fixed bone lengths.
Like our method, IK and its variants can also handle arbitrary open-set control tasks, so we compare with IK and IK+Reg. in all quantitative and qualitative experiments.

\noindent\textbf{Inpainting-based methods.}
MDM Edit and PriorMDM finetuned control are inpainting-based methods. They support motion control tasks by assigning exact joint trajectories. However, they cannot natively handle tasks described by constraints, especially, inequality constraints. Moreover, PriorMDM needs to finetune the network for controlling a specified joint and only finetuned models for hand, foot and root trajectories are provided \cite{shafir2023human}. For the above reasons, we only compare with MDM Edit on trajectory control-based tasks, \textit{i.e.}, Task HSI-1, Task GEO-1 and Task HOI-1, and we compare with PriorMDM on Task GEO-1 and Task HOI-1, which only involves hand trajectory control.

In their original papers, MDM Edit and PriorMDM finetuned control only support inpainting with root trajectories and valid local joint positions. Since the constraints for tasks defined in Section 4 are majorly represented in global coordinates, we adapt MDM Edit and PriorMDM control to handle control signals in global positions. Specifically, we first generate a sample and take its root trajectory. We then use ad-hoc tricks to generate a trajectory for the control joint in global positions that satisfies the given constraint and further convert it to local positions given the root trajectory. Finally we inpaint both the root trajectory and the local trajectory of the control joint. Similar to IK, as recovering from local joint positions yields invalid bone lengths (see Table 3 in the main paper), we obtain the global motion from joint rotations using a human skeleton template with fixed bone lengths. Also, for PriorMDM we use model blending \cite{shafir2023human} for inpainting both root trajectory and control joint trajectory. 

\red{The ad-hoc tricks are designed as follows: for Task HSI-1 and HOI-1, we directly set the key-frame positions with the required constraint. For Task GEO-1, we project the generated hand trajectory onto the given plane to obtain the new hand trajectory in the global positions.}

\subsection*{A.3 Implementation Details}

Following unconstrained MDM, we also recover global positions from local joint positions.
\red{Since the error function for each task may vary, while optimizing with learning rate 0.005 and 100 optimization steps generally works well for a majority of tasks, we may also increase the initial learning rate to up to 0.05 for faster convergence in some cases.} \revise{Besides, we may add regularization term using absolute position constraint to preserve desired motion characteristics for root trajectory or body parts in some cases.}

\noindent\textbf{Constraint relaxation.}
\revise{We only apply constraint relaxation on Task GEO-1, Task GEO-2 and Task HOI-1, which involves absolute position constraints of point, line and plane. It takes advantage of translation invariance of motion for fast convergence and compensates for the limited horizontal space coverage of root trajectories in the original motion prior.}
For Task GEO-1, we relax the plane constraint by fitting the generated hand trajectory on an optimal vertical plane. For Task GEO-2, we relax the line constraint by fitting the foot trajectories on an optimal line. For Task HOI-1, we relax the required beginning and end points $A, B$ to fall on the line connecting the beginning and end points generated by the model $\hat A, \hat B$ and keep their middle points the same, \textit{i.e.}, 
$A_{relax}=P+ \frac{\hat A-P}{|\hat A - P|} \frac{|A-B|}{2}, B_{relax}=P+ \frac{\hat B-P}{|\hat B - P|} \frac{|A-B|}{2}$, where $P=(\hat A + \hat B)/2$.

In practice, we update the constraint using the aforementioned relaxation strategy every $K$ steps and minimize the constraint error for $x$ using the updated constraints. In this way the whole optimization process can be implemented as relax-and-minimize loops.
For a fair comparison, IK and IK+Reg. also use constraint relaxation for experiments in Table 2 in the main paper.

\setlength{\tabcolsep}{10pt}
\begin{table}[t]
    \centering
    \small
    \begin{tabular}{lccc} 
    \toprule
    \multicolumn{4}{c}{Task GEO-1: hand touching wall} \\
    \toprule
   Method            & Foot Skate  & Max Acc.  & C.Err. \\  
   \midrule
   IK w/o relax.    & 0.375 & 0.209 & 0.210    \\   
   IK w/ relax.     & 0.187 & 0.147 & \textbf{0.010}  \\ 
    \midrule
   Ours w/o relax.  & 0.094 & 0.129 & 0.118    \\   
   Ours w/ relax.   & 0.110 & 0.104 & \textbf{0.023}  \\ 
  \bottomrule
  \toprule
    \multicolumn{4}{c}{Task HOI-1: moving object} \\
    \toprule
   Method            & Foot Skate  & Max Acc.  & C.Err. \\  
   \midrule
   Ours w/o relax.  & 0.078       & 0.077     & 0.069   \\ 
   Ours w/ relax.   & 0.114       & 0.068     & \textbf{0.028}  \\ 
  \bottomrule
    \end{tabular}
    \caption{Effect of constraint relaxation. Constraint relaxation helps better reach constraints related to horizontal positions for optimization-based methods.}
    \label{tab:relax_appendix}
\end{table}
\setlength{\tabcolsep}{1.4pt}

\setlength{\tabcolsep}{4pt}
\begin{table}[t]
    \centering
    \small
    \begin{tabular}{lcccc} 
    \toprule
    \multicolumn{5}{c}{Task HSI-1: head height constraint} \\
    \toprule
   Method & Foot Skate  & Diversity & FID & C.Err. \\  
   \midrule
   MDM (Unconstrained) &            0.086  & 9.656  & 0.545 & 0.118 \\
   Ours ($N_S=1$) & 0.075	& 9.611	& 0.556	& 0.012  \\ 
   Ours ($N_S=5$) & 0.072	& 9.422	& 0.648	& \textbf{0.002}  \\ 
  \bottomrule
    \end{tabular}
    \caption{Effect of initial point search. $N_S$ denotes the number of searches. Using a random initial point search leads to significantly smaller constraint error. It provides a solution for generating motions that better adhere to the given constraint.}
    \label{tab:initial_point_search}
\end{table}
\setlength{\tabcolsep}{1.4pt}

\subsection*{A.4 Experiment Details for Bone Length Preserving} 
We provide more experimental details for Table 3 in the main paper. For the generated motions in Task HSI-1 in Table 1, we investigate the neck length (bone length between joint 12 and 15) at the key-frames where the head height constraint is imposed. We empirically set a range between 0.08-0.025 and 0.08+0.025, and the neck length which falls outside this range is considered as incorrect bone length. The bone length incorrect ratio is defined as the ratio of key-frames with incorrect neck lengths in all the generated key-frames. We find that unconstrained MDM and our method have low incorrect ratio even if we directly recover global positions from local joint positions. However, if we recover motions generated by MDM Edit from local joint positions, the incorrect ratio becomes very large, indicating that a great percentage of the generated samples are of invalid human layouts. \red{For this reason, we choose to recover global motion from joint rotations for inpainting-based methods MDM Edit and PriorMDM}.

\subsection*{A.5 Additional Analysis}

\noindent\textbf{Effect of constraint relaxation.} As in Table \ref{tab:relax_appendix}, the constraint relaxation strategy significantly reduces the constraint error for goal reaching tasks on the horizontal plane, such as task \emph{hand touching wall} and \emph{moving object}. While the constraints are better satisfied, we observe slight decrease in motion quality, which is indicated by Foot Skate. Also, it is shown that the constraint relaxation is a general optimization strategy since there is a significant decrease in the constraint error for IK as well.

\noindent\textbf{Effect of initial point search.}
The initial noise $z$ may affect the final constraint error if the initialized motion is too far away from reaching the constraints. A straightforward way would be to sample random noise $z$ in several runs and pick the result with the smallest constraint error.
We conduct experiment on Task HSI-1 using the same setting as Table 1 in the main paper and compare the results of $N_S=1$ and $N_S=5$. Here $N_S$ denotes the number of initial point searches. The results are shown in Table \ref{tab:initial_point_search}.
We observe that using a random initial point search leads to significantly smaller constraint error but at the cost of diversity and FID scores. It provides a solution for generating motions that better adhere to the given constraint.

\noindent\textbf{Diversity of generated motions.}
By optimizing the latent vector of generated motions to conform to the motion prior, our method can generate diverse motions under the same constraint. For example, in the task of \emph{left hand always touching head}, apart from single hand touching the face, we observe that constraining only one hand can also give rise to the touching of another hand. (see Fig. 1 and Fig. 4 in the main paper).

\section*{B. Details for Motion Programming by LLM}
Our programmable motion generation framework also makes automatic programming possible with the aid of large language models (LLM).
As in Fig. \ref{fig:llm}, in order to generate code for the error function $F$, we first feed instructions to GPT \cite{brown2020language} with the rules and ingredients for motion programming, e.g., input arguments and functions in the atomic constraint library. After that, one can feed the textual description for an arbitrary open-set motion control task to GPT. In Fig. \ref{fig:llm} we show the textual input fed to GPT as well as the raw code output by GPT for Task GEO-1, HSI-3 and HSI-4 in the main paper. We observe that an LLM can pick correct atomic constraints, logical operations (e.g. ``$>$'', ``$<$''), and procedural operations (e.g. if-else clauses) for given tasks. Note that the code blocks labeled with GPT markers for Task GEO-1 and Task HSI-4 in Fig. 4 in the main paper are slightly modified in the coding style to make them consistent with other manually written code, without changing the code logic. 

\setlength{\tabcolsep}{3pt}
\begin{table}[t]
    \centering
    \small
    \begin{tabular}{lc} 
    \toprule
    Evaluation tasks \\
    \toprule
    \emph{walking with hand always touching face.} & \checkmark \\
    \emph{walking inside a square.} & \checkmark \\
    \emph{carrying a ball.} \\
    \emph{carrying a heavy ball.} \\
    \emph{walking with feet on a straight line.} &\checkmark \\
    \emph{walking with hand touching a wall.} &\checkmark \\
    \emph{walking in a gap between two walls.} &\checkmark \\
    \emph{walking to avoid an overhead barrier.} &\checkmark \\
    \emph{picking object from A to B.} &\checkmark \\
    \emph{walking with velocity constraint on three frames.} &\checkmark \\
    \emph{standing and keeping balanced with single foot.}  \\
    \emph{walking with head height constraint on three frames.} &\checkmark \\
    \emph{lying on a bed.} &\checkmark \\
    \emph{sitting on a chair.}  \\
    \emph{kicking a ball in the last frame.} &\checkmark \\
    \emph{walking with both hands in contact.} &\checkmark \\
    \emph{jumping over a barrier.} \\
    \emph{pointing to a direction with left arm.} &\checkmark \\
    \emph{dancing with specified velocity magnitude on three frames.} &\checkmark \\
    \emph{twisting for two circles.} \\
  \bottomrule
    \end{tabular}
    \caption{Evaluation on motion programming by LLM. Tasks that are successfully handled by LLM are labeled with \checkmark.}
    \label{tab:evaltasks}
\end{table}
\setlength{\tabcolsep}{1.4pt}

\noindent\textbf{More evaluation.}
As in Table \ref{tab:evaltasks}, we design 20 unique tasks (including those presented in the main paper), and evaluate the success rate of LLM programming via comparing to manual programming. With little prompt engineering, the success rate turns out to be 14/20. \revise{In failure cases, it typically picks incorrect inequality logical operations, or provides excessive and incorrect physical constraints.}
Nevertheless, we find that LLM comes up with novel constraints beyond manual programming, e.g. tilt angle constraint for the action \emph{balancing}.

\section*{C. Discussion and Limitations}

\revise{\noindent\textbf{Sources of error.} 
As we propose a general framework for open-set motion control tasks, the performance of individual modules can be further improved.
First, we observe some unrealistic poses and motion artifacts in our generated motions. 
Since the FID score shows that our results have similar quality to unconstrained MDM (See Table 1 in the main paper), a possible solution is to enlarge the pre-trained model together with more training data.}
\revise{Also, for complex tasks, either an end user or an LLM may have difficulty of crafting detailed and appropriate constraints, which is likely to lead to unnatural motions.}
Second, the constraint error sometimes remains big compared to IK, for example, for the unseen Task HSI-2. 
Although it is reasonable that IK directly optimizes on motion $x$ and thus has less difficulty for reaching the constraint, we will further investigate better optimization approaches to solve this issue. 
Possible solutions include (1) combining optimized and IK-based motion in the denoising process, (2) relaxing on the parameters of the generation model and involving it in the optimization process like \cite{pan2021exploiting}, and (3) searching for more suitable optimizers.

Moreover, the action semantics for the generated motion is observed to change slightly in the experiments, e.g. for Task HSI-1. This calls for more suitable generation models and optimization strategies that can better adhere to the text condition.

\noindent\textbf{Coverage of the proposed constraint library.} We examine the coverage of our proposed library for daily motions on BABEL-120 dataset \cite{punnakkal2021babel}. We find that nearly 16\% of the actions involve periodic, rotational or symmetric movement, whose control is not directly supported by our library. \revise{We plan to further add frequency-domain, rotational and symmetric constraints into our library.}

\noindent\textbf{Comparison with reinforcement learning and trajectory optimization.}
\revise{We note that RL-based \cite{peng2017deeploco, peng2018deepmimic, kwiatkowski2022survey} and trajectory optimization approaches \cite{al2012trajectory} also build compositional reward or goal functions for specialized motion control tasks, and we here provide a discussion for these approaches:}
(1) Based on our experiments, the error function design in this work is not as difficult as reward design in reinforcement learning (RL), not only because the error function only handles the constraints, but the optimization in latent space is easier to converge than RL training, since the pre-trained model already provides a neat and smooth manifold as the optimization space.
(2) The pre-trained generation model is easier to accommodate more motion skills and scale up with more data. This is the main consideration for us towards solving open-set tasks. RL usually requires specific design to support diverse tasks \cite{won2020scalable}.
(3) It is easier than RL to control the semantics via text condition.
(4) RL and ours are complementary. RL has better physics-grounded qualities.
(5) Compared to trajectory optimization, optimizing latent code better preserves semantics imposed by text condition. Besides, optimizing latent code may be more advantageous for composing novel types of actions since it acts like semantic interpolation in the data distribution. Trajectory optimization normally optimizes on one reference motion \cite{liu2018learning, gartner2022trajectory}.

\noindent\textbf{Time performance.}
Currently it costs a few minutes for each customized task, but is still much better than previous works that require collecting new data and training new networks.
We have not focused a lot on improving optimization efficiency in this work, which might be a direction in the follow-up works.
Although not applicable to real-time generation, it is suitable for off-line content creation due to its high customizability.